\documentclass[runningheads]{llncs}
\usepackage[T1]{fontenc}
\usepackage{graphicx}
\usepackage{hyperref}
\usepackage{url}
\usepackage[utf8]{inputenc} 
\usepackage[T1]{fontenc}    
\usepackage{hyperref}       
\usepackage{url}            
\usepackage{booktabs}       
\usepackage{amsfonts}       
\usepackage{nicefrac}       
\usepackage{microtype}      
\usepackage{xcolor}         
\usepackage{float} 
\usepackage{wrapfig}
\usepackage{amsfonts}
\usepackage{multirow}
\usepackage{subcaption}
\usepackage{ragged2e}
\usepackage{amsmath}
\usepackage{amssymb}
\usepackage{thmtools} 
\usepackage{makecell}
\usepackage[table]{xcolor}
\usepackage{ulem}
\newcommand{\citep}[1]{\cite{#1}}
\newcommand{\citet}[1]{\cite{#1}}

\newcommand{\cyl}[1]{{\color{black}#1}}

\newcommand{\yhs}[1]{{\color{black}#1}}
\newcommand{\chris}[1]{{\color{black}#1}}

\begin{document}
\title{Dynamic Context Adapters: Efficiently Infusing History into Vision-and-Language Models}
\titlerunning{Dynamic Context Adapters}
%
\author{Yuhang Song\inst{1,2} \and
Bor-Jiun Lin\inst{3} \and
Jiaxu Liu\inst{3} \and
Te-Chuan Chiu\inst{2} \and
Anh Nguyen\inst{1} \and
Chun-Yi Lee\inst{4}}
\authorrunning{Y. Song et al.}
%
\institute{Department of Computer Science, University of Liverpool \\
\email{sgyson10,Jiaxu.Liu,Anh.Nguyen@liverpool.ac.uk} \and
Department of Computer Science, National Tsinghua University \\
\email{theochiu@cs.nthu.edu.tw} \\ \and
Imperial College London \\
\email{j.liu2@imperial.ac.uk} \\ \and
Department of Computer Science and Information Engineering \& \& NTU Artificial Intelligence Center of Research Excellence (AI-CoRE), National Taiwan University  \\
\email{crlc112358@gmail.com}, \email{cylee@csie.ntu.edu.tw} \\
}
\maketitle              
\begin{abstract}\vspace{-2em}
Historical context integration presents a fundamental challenge for Vision-Language Models (VLMs) in sequential decision-making tasks. Current VLMs process visual inputs independently, which creates critical limitations for downstream applications that require temporal understanding. Direct incorporation of historical frames into Transformer inputs produces quadratic attention complexity and excessive memory consumption. Existing approaches suffer from significant drawbacks: computational inflation or substantial information loss through temporal compression. To address these challenges, we introduce Dynamic Context Adapter (DCA), a novel context injection approach for pretrained VLMs. Our method employs fixed-size, dynamically compressed memory to preserve historical semantics without frame concatenation. DCA bridges static VLMs and recurrent policies and enables memory capabilities in pretrained models while maintaining computational efficiency. DCA achieves over $25\%$ reduction in attention FLOPs and $13\%$ memory savings while improving performance on long-horizon tasks.

\end{abstract}

\section{Introduction} 
\vspace{-1em}
\label{Sec:intro}
\begin{figure}[htbp]
\centering
\resizebox{0.85\textwidth}{!}{\includegraphics{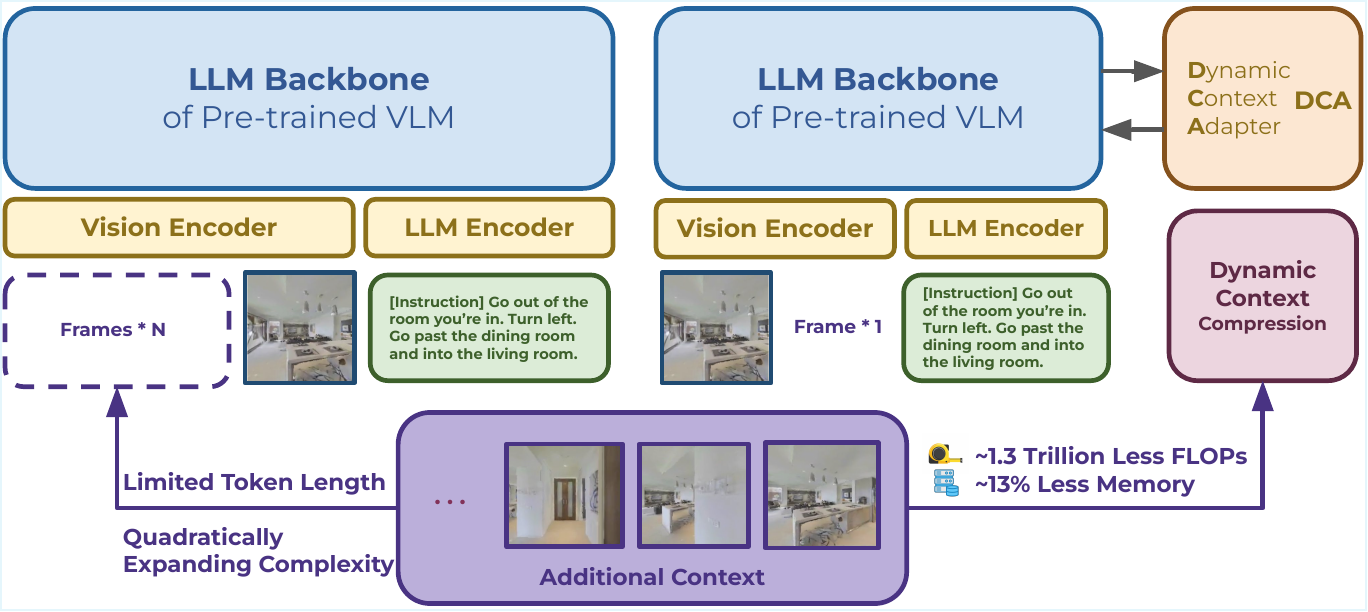}}
\caption{\textbf{Context Concatenation vs. Dynamic Context Adaptation.}
\textbf{(Left)} Traditional concatenation appends historical frames to current input, producing quadratic computational complexity and token length constraints. \textbf{(Right)} Our proposed DCA method decouples historical context from the LLM backbone through lightweight adapters. The Dynamic Context Compression module processes historical frames and distributes compressed representations across multiple VLM layers, maintaining constant input length while achieving about $\sim1.3$ trillion fewer FLOPs and 13\% memory reduction.
} \vspace{-1em}
\label{intro_overview} \vspace{-1.3em} 
\end{figure}

\cyl{Sequential decision-making in partially observable environments demands agents to integrate rich historical context across extended temporal horizons. In Vision-and-Language Navigation (VLN), agents must synthesize information from multiple past observations to navigate complex multi-room environments where current visual input alone provides insufficient context for decision-making. Long-horizon instructions such as \emph{``pass through the bedroom, locate the study, and pick up the book on the desk"} require agents to chain subgoals while preserving spatial dependencies across rooms and corridors. Under partial observability, an agent's onboard camera captures only a limited field of view at each timestep, making historical context essential for inferring occluded landmarks, retracing steps, and maintaining spatial awareness. While Transformer-based Vision-Language Models (VLMs)~\citep{vaswani2017attention,kim2021vilt,liu2023visual,li2023blip,alayrac2022flamingo,bai2023qwen,chen2024far} have achieved remarkable success in single-frame visual reasoning tasks such as Image Captioning~\citep{hossain2019comprehensive} and Visual Question Answering~\citep{antol2015vqa}, their adaptation to sequential tasks reveals fundamental limitations. Recent approaches in Vision-and-Language Action~\citep{ma2024survey} and Navigation~\citep{wu2024vision} tasks, including OpenVLA~\citep{kimopenvla}, RT-1~\citep{brohan2022rt}, RT-2~\citep{brohan2023rt}, Navid~\citep{zhang2024navid}, UniNavid~\citep{zhang2024uni}, and NavGPT2~\citep{zhou2024navgpt2}, have demonstrated the potential of VLMs in embodied scenarios. However, these methods struggle to efficiently integrate the extensive historical visual context necessary for long-horizon reasoning.}

Existing strategies for integrating historical context into VLM backbones can be grouped into three main categories.  
(1) Token concatenation approaches are widely used in integrating historical context in Transformer-based models~\citep{zhang2023video,zhang2024navid,hamt,Guhur_2021_ICCV,Lin_2023_ICCV}. (2) Recurrent compression methods employ RNNs or LSTMs to compress the entire frame history into a single state vector~\citep{krantz2020beyond,abs-2011-13922}. (3) Previous studies also evaluate methods that maintain an external mapping and memory frameworks by external topological or semantic maps~\citep{mapnav,bev,etp,weakly_super}. Although these methods demonstrate \cyl{effective} performance, all three classes face certain limitations when applied to large-scale pretrained VLMs in environments \cyl{that require} context. Token concatenation disrupts the downstream token order and floods the model with redundant information. Recurrent compression lacks the capacity to represent fine temporal structure, leading to information loss over extended sequences. External memory methods depend on manually constructed maps that may not generalize across different environments. 
\cyl{These limitations collectively highlight the necessity for a more efficient and effective methodology to integrate long historical context into pretrained VLMs.}

\cyl{Motivated by these challenges, in this study, we focus on eliminating memory bottlenecks and reducing computational complexity while preserving the original model architecture to maintain effectiveness. We draw insight from recent advances in parameter-efficient fine-tuning (PEFT) for large language models (LLMs)~\citep{hu2022lora,zhang2023llama,kim2025fine}, which insert small trainable modules into frozen backbones with minimal overhead. This motivates our investigation of whether a similarly lightweight adapter paradigm can fuse rich historical visual information into VLMs while preserving their efficiency and pretrained knowledge. To this end, we propose the \textbf{D}ynamic \textbf{C}ontext \textbf{A}dapter (DCA), which compresses arbitrary sequences of past frame embeddings into a fixed set of learnable context vectors. DCA eliminates the memory bottleneck associated with naive token concatenation while capturing rich temporal semantics. To enable the model to consult its memory at every depth without altering original parameters or structure, these compressed representations are adapted to the LLM outputs through lightweight adapter modules and injected into each layer of the pretrained VLM. Our method delivers three key advantages. \textbf{First}, DCA ensures computational efficiency by maintaining constant input token length regardless of past frame quantity, achieving linear complexity growth with extended context. \textbf{Second}, DCA preserves fine-grained contextual details through dynamic compression of critical information into fixed context vectors. This approach avoids temporal detail loss common in recurrent models while removing redundant features. \textbf{Third}, DCA retains the original input and fully preserves the priors of the pretrained VLM, which enables maintaining its learned knowledge to the greatest extent possible.}

\cyl{We evaluate DCA on the standard navigation benchmark and compare it against both RGB-only baselines and existing context-integration approaches. Our findings suggest that DCA resolves the core tension between capturing comprehensive visual history and maintaining computational tractability. The experimental results demonstrate that DCA matches or exceeds prior methods in Success Rate while reducing attention FLOPs by over $25\%$ and cutting peak memory consumption by $15\%$ on long-horizon VLN tasks. Our contributions can be summarized as follows:} (1) We introduce DCA, an efficient and lightweight framework that addresses the core limitation of VLMs in sequential tasks by enabling dynamic compression and integration of historical visual context without disrupting pretrained model architecture or inflating input sequences. (2) We demonstrate that DCA overcomes the fundamental challenge of information loss in recurrent-based approaches and memory explosion issues in concatenation methods, enabling VLMs to maintain rich temporal understanding across extended navigation episodes. (3) We validate that DCA enables effective utilization of historical context for long-horizon reasoning in partially observable environments, achieving superior navigation performance.

    

\section{Related Works} \label{Sec:Related_works}
\vspace{-0.5em}
\textbf{Pretrained Vision-Language Models.} Large-scale VLMs~\citep{kim2021vilt,grattafiori2024llama,touvron2023llama,karamcheti2024prismatic,li2024llama,bai2023qwen,chen2024far} have achieved impressive multimodal general-purpose reasoning capabilities. For example, ViLT~\citep{kim2021vilt} introduced a minimalist vision-language Transformer that forgoes region-based visual features for end-to-end image-text encoding. Likewise, LLaVA~\citep{liu2023visual} fine-tunes a pre-trained vision encoder together with a fine-tuned version of LLaMA~\citep{genai2023llama} using GPT-4~\citep{achiam2023gpt} generated instruction follow-up data, producing a powerful multimodal assistant capable of open-ended visual dialogue. These models are typically designed for textual modality or static image-text pairs, and do not accommodate video or historical visual contexts essential for navigation and temporally extended reasoning tasks. LLaMA-VID~\citep{li2024llama} extends the LLaMA~\citep{genai2023llama} for video-text tasks but primarily addresses the problem with naive inefficient token concatenation.

\textbf{Navigation with Pretrained Large Models.} Several recent works have explored applying foundation models to embodied VLN tasks. Intuitive approaches involve directly leveraging pretrained large language models as planners~\citep{xuvision,shah2023lm,zhou2024navgpt,long2024discuss,chen2024mapgpt,chen2025affordances,weerakoon2024behav}, while other groups of works have shown great success in incorporating VLMs as navigation backbones~\citep{zhou2024navgpt2,lin2025navcot,pan2024langnav,liu2025spatialcot,zheng2024towards,zhang2024uni}. NaVid~\citep{zhang2024navid} fine-tunes a video-based VLM backbone to predict next-step actions by concatenating raw frame tokens, including both current and historical observations. The following work Uni-NaVid~\citep{zhang2024uni} unifies multiple navigation tasks in a video-based VLM backbone, processing long video streams end-to-end. ~\cite{zhou2024navgpt2} augments \chris{LLMs with policy networks for VLN by input concatenation. While leveraging powerful pre-trained representations, they suffer from quadratic scaling with frame concatenation and lack mechanisms to distill and recall prior observations. DCA uses lightweight adapters to effectively decouple history context from LLM input, maintaining constant-length inputs while efficiently retrieving relevant historical information across layers.}

\textbf{Historical Context in Navigation.}  \cyl{Traditional recurrent models maintained implicit memory via LSTM or GRU hidden states that carry over past perceptions~\citep{anderson2018vision,tan2019learning,fried2018speaker,krantz2020beyond,pmlr-v222-song24a,abs-2011-13922,sim2sim,MEM}, but more recent approaches use Transformer-based architectures to capture longer-range dependencies~\citep{alignvln,zhang2024uni,song2024fine,Lin_2023_ICCV,Guhur_2021_ICCV,hamt,zhang2024navid}. These methods integrate historical context by either maintaining recurrent hidden states or concatenating history frames as additional input during prediction, which may result in information loss. Other works have proposed building structured memory representations of the environment~\citep{birdeye,bev,etp,Evolving,Wang_2023_ICCV,nerf,topologic,chen2022thinkglobalactlocal,weakly_super,mapnav}. Our work uses Transformer-based pretrained VLMs as backbones, but instead of adding additional input tokens, we introduce an efficient method to adapt context into LLM layers. Previous works were designed for static vision-language alignment tasks like few-shot prompting or pretraining with fixed image-text pairs~\citep{manas2022mapl,clip}. In contrast to these prior methods that perform one-time modality bridging, our model operates within a Partially Observable Markov Decision Process (POMDP) and continuously compress expanding history.}


\vspace{-1em}

\section{Efficient History Context Adaptation \cyl{Methodology} 
} \label{Sec:method}
\vspace{-0.5em}

\cyl{\textbf{Efficient VLM Architecture Selection.} To maximize efficiency while demonstrating effectiveness, we employ a compact pretrained VLM backbone. Following recent advances in efficient VLM deployment~\citep{zhang2024navid,zhang2024uni,mapnav,kimopenvla}, we adopt PrismaticVLM~\citep{karamcheti2024prismatic} as our foundation. We select the \emph{phi-2+3b} variant with only \emph{3B} parameters, which incorporates a ViT-based CLIP~\citep{vit,clip} visual encoder, a lightweight \emph{Phi-2}~\citep{unknown} language model, and multi-layer cross-modal projection. This architecture choice demonstrates that our efficiency gains extend beyond large-scale VLM models.}

\cyl{\textbf{Efficient Context Processing Pipeline.} Given visual observations $X$, we encode each frame into visual tokens and project them into a shared embedding space with language tokens. This process yields $X^\prime = \{X^\prime_{1:t-1}, X^\prime_t\}$. Instructions $L_t$ are tokenized to produce $L^\prime_t$. For action prediction at timestep $t$, we process current frame tokens $X^\prime_t$ and instruction tokens $L^\prime_t$ through the LLM while utilizing encoded historical frames $X^\prime_{1:t-1}$ as inputs for efficient contextual embedding adaptation in LLM layers. }

\begin{figure} 
\centering
\includegraphics[width=0.85\textwidth]{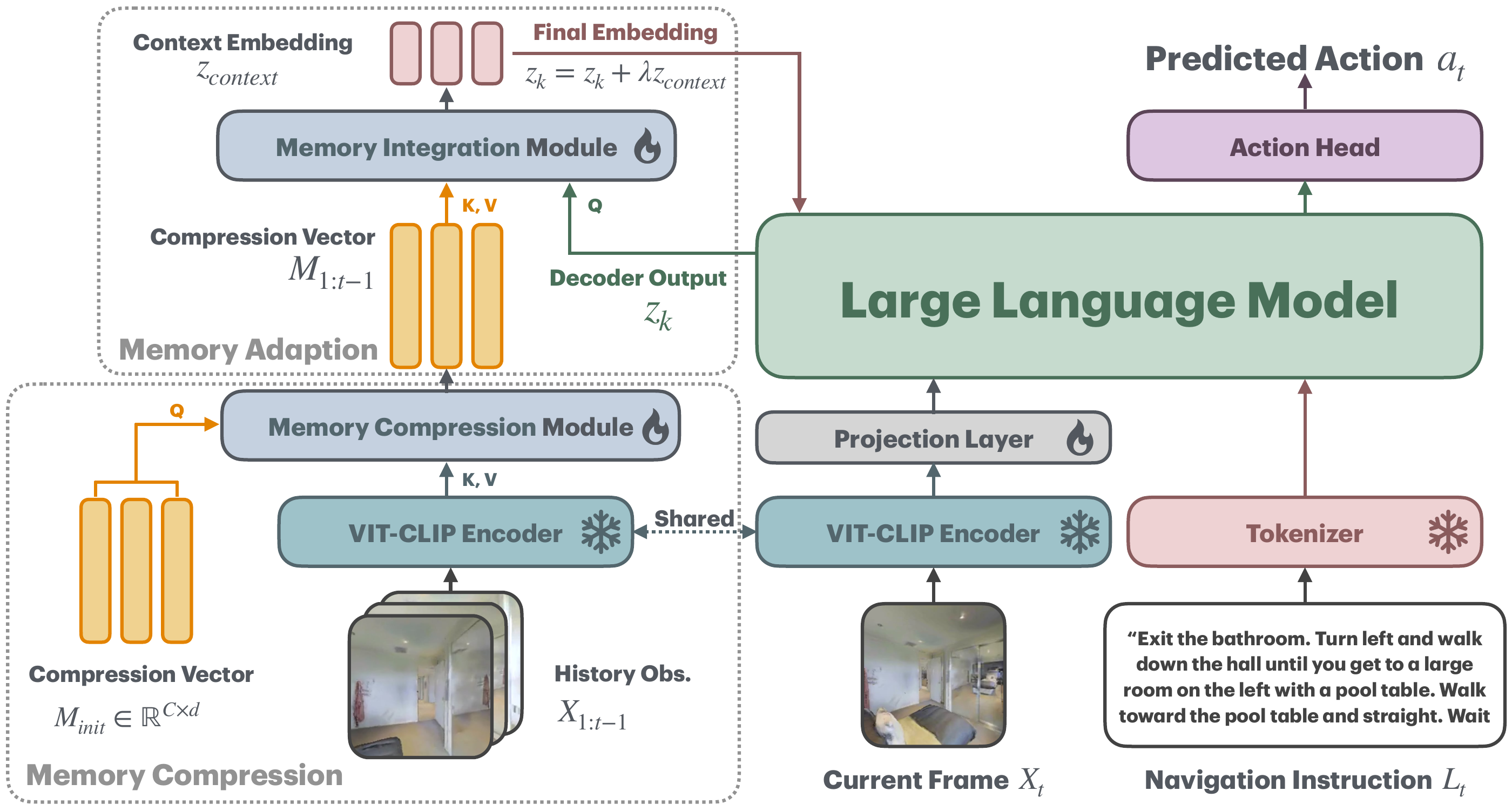}
\caption{
Model Architecture Overview. In each timestep t, the model receives inputs, including initial compression vector, history observations, current observations, and navigation instruction. We compress the historical context through a \emph{Memory Compressing Module}, pass it to the \emph{Memory Integration Module}, and adapt the resulting memory into the layer outputs of the LLM backbone.
}
\label{model_architecture} \vspace{-1.1em}
\end{figure}

\vspace{-1.5em}

\vspace{-0.5em}
\subsection{Dynamic Context Adaptation \cyl{for Efficient Historical Integration}} 
\vspace{-0.5em}

\cyl{Our DCA addresses the computational bottleneck of historical context integration through a two-stage architectural design that maintains linear complexity while preserving temporal richness. The core innovation lies in decoupling historical context processing from the main VLM backbone, which enables efficient memory management without sacrificing representational capacity. Given that navigation environments constitute POMDPs, agents must integrate previous observations for informed decision-making. However, naive concatenation of past tokens causes input sequence explosion and incurs super-linear self-attention costs (detailed in supplementary). DCA resolves this efficiency-accuracy trade-off by dynamically compressing historical context into a compact set of learnable memory vectors that adapt into LLM layers while preserving upstream pretrained semantics.}

\cyl{Fig.~\ref{model_architecture} illustrates our efficient dual-pipeline architecture. The standard VLM pathway processes $X_t$ through a shared visual encoder and tokenizes instruction $L_t$ via the Phi tokenizer. Both inputs pass through the pretrained VLM to produce decoder embedding $\mathbf{z}_t$, which an action head decodes into next-step action $a_t$ following standard next-token prediction. The efficiency-focused context adaptation pathway operates in parallel: a fixed-size learnable compression vector $\mathbf{M}_{\mathrm{init}}$ queries past embeddings $X_{1:t-1}$ through our \emph{Memory Compression Module}, producing compressed memory $\mathbf{M}_{1:t-1}$. Our \emph{Memory Integration Module} then attends over $\mathbf{M}_{1:t-1}$ with current decoder queries to extract context-enhanced outputs that can adapt into LLM layers without inflating input sequences.}

\vspace{-0.5em}
\subsubsection{\cyl{Efficient} Dynamic Context Vector Compressing}
\label{sec:compression}
\vspace{-0.5em}

\cyl{Our compression strategy achieves computational efficiency by transforming variable-length historical sequences into fixed-size representations while preserving critical temporal information. This design eliminates quadratic scaling in concatenation methods and enables practical deployment in resource-constrained scenarios. We initialize a learnable compression vector $\mathbf{M}_{\mathrm{init}} = \texttt{nn.Embedding}(C,d).\texttt{weight} \in \mathbb{R}^{C \times d}$ for each timestep $t$, where $C$ denotes memory token count and $d$ represents embedding dimension. Historical frames $X_{1:t-1}$ are encoded via the vision encoder~\citep{vit,clip} and concatenated to form encoded features $\mathbf{F}_{1:t-1} = \|_{t=1}^{t-1} \text{ViT-CLIP}(\mathbf{X}_i) \in \mathbb{R}^{(t-1) \times P \times d}$, where $P$ denotes image patch count. To reduce spatial redundancy, we apply grid pooling operator $\mathcal{G}:\mathbb{R}^{P \times d}\to\mathbb{R}^{p \times d}$ (with $p \ll P$) following established practices~\citep{zhang2024navid,li2024llama}. This yields $\mathbf{F}_{1:t-1} = \mathcal{G}(\mathbf{F}_{1:t-1}) \in \mathbb{R}^{(t-1) \times p \times d}$. Our \emph{Memory Compression Module} employs multi-layer cross-attention between $M_{\text{init}}$ and pooled features. We project $M_{\text{init}}$ as queries and history features as keys and values: $Q_M = M_{\text{init}} W_Q$, $K_F = \mathbf{F}_{1:t-1} W_K$, $V_F = \mathbf{F}_{1:t-1} W_V$. The compressed computation achieves $O(C \cdot p)$ complexity:}
\begin{equation}\label{eq:compressed memory vector}
M_{1:t-1} = S_{\text{cps}} V_F \in \mathbb{R}^{C \times d}, \quad \text{where} \quad S_{\text{cps}} = \text{Softmax}(Q_M K_F^T) \in \mathbb{R}^{C \times p}
\end{equation}

\vspace{-0.5em}
\subsubsection{
\cyl{Efficient Context Adaptation for LLM Integration}
}
\vspace{-0.5em}
\cyl{Our context adaptation integrates compressed historical information directly into LLM layers without inflating input sequences or disrupting the original architecture. This approach maintains constant computational overhead regardless of history length while enabling multi-layer conditioning that enhances temporal understanding. The integration process operates on standard encoder-only multi-layer language models. For each layer $k$ with input $z_{k-1} \in \mathbb{R}^{S \times d}$, where $S$ represents sequence length, the standard layer output $z_{k}$ is formulated as:}
\begin{equation}\label{eq:original output}
\begin{split}
z_k &= \mathrm{Atten}\bigl(Q_{k-1},\,K_{k-1},\,V_{k-1}\bigr), \\
Q_{k-1},K_{k-1},V_{k-1} &= z_{k-1}W_Q^{k-1},z_{k-1}W_K^{k-1},z_{k-1}W_V^{k-1}.
\end{split}
\end{equation}
\cyl{Our \emph{Memory Integration Module} enables efficient historical context adaptation into each Transformer layer. This module integrates the compressed context vector from Eq.~(\ref{eq:compressed memory vector}) through lightweight cross-attention that maintains linear complexity. The module projects compressed historical context into key-value representations: $K_M = M_{1:t-1} W_K^M$ and $V_M = M_{1:t-1} W_V^M$. The context-enhanced output computation achieves efficiency by attending the original layer output $z_{k}$ to compressed historical vectors rather than processing full sequence history, which can be expressed as follows:}
\begin{equation}\label{eq:context-enhanced output}
z_{k}^{\text{context}} = S_{\text{intg}} V_M, \quad \text{where} \quad S_{\text{intg}} = \text{Softmax}(Q_{k-1} K_M^T),
\end{equation}
\cyl{where $S_{\text{intg}}$ denotes the attention score of the integration module. The final layer output combines the context-enhanced representation with the original output through learnable scalar weighting as:}
\begin{equation}\label{lambda}
z_{k+1} \leftarrow z_{k+1} + \lambda z_{k+1}^{\text{context}}.
\end{equation}
\cyl{This design maintains computational efficiency by processing only $C$ compressed memory tokens per layer rather than the full history sequence of length $t$, achieving favorable $O(S \cdot C)$ for efficient context integration compared to the prohibitive $O(S \cdot t \cdot p)$ for the naive concatenation approaches.}

\vspace{-1.0em}
\section{Experimental Results} \label{Sec:exp}


\vspace{-0.5em}
\subsection{Experimental Setup}
\vspace{-0.5em}

\begin{wraptable}{r}{0.6\textwidth} 
\centering
\vspace{-1.5em} 
\caption{Inference Throughput Comparisons.}
\label{tab:efficiency}
\scriptsize
\rowcolors{2}{cyan!20}{white}
\setlength{\tabcolsep}{3pt} 
\begin{tabular}{l c c c c}
\Xhline{1.2pt}
\makecell[l]{Method\\(Input: RGB)} & 
\# Params & 
\makecell{Step Inf.\\Time} & 
\makecell{FLOPs\\(T)} & 
\makecell{Mem.\\Peak\\(GB)} \\
\midrule
Navid-IL & 7B & 2.86 & 4.89 & 48.61 \\
No-Adapt & \textbf{3B} & 3.21 & 4.77 & 37.84 \\
Recurrent-Adapt & \textbf{3B} & \textbf{2.50} & \textbf{4.14} & 35.65 \\
DCA (Ours) & \textbf{3B} & 2.71 & 4.23 & \textbf{34.31} \\ 
\Xhline{1.5pt}
\end{tabular}
\vspace{-2em}
\end{wraptable}

\textbf{Baselines.} \cyl{For a fair comparison, we evaluate methods that implement end-to-end learning with low-level action primitives in the VLN-CE environments. \textbf{(1) Seq2Seq~\citep{krantz2020beyond}}: A recurrent sequence-to-sequence architecture that directly maps RGBD observations to navigation actions. The RGB-Seq2Seq variant processes RGB inputs exclusively. \textbf{(2) CMA~\citep{krantz2020beyond}}: Implements cross-modal attention between instructions and RGBD observations for action prediction. Note that RGB-CMA denotes the RGB-only configuration. \textbf{(3) NaVid~\citep{zhang2024navid}}: Employs a frozen VLM backbone to formulate navigation as next-token prediction over RGB sequences. This method concatenates historical observations as additional language tokens and applies auxiliary training objectives. NaVid-IL represents the imitation learning configuration. For efficiency experiments, we establish two controlled baselines that share our VLM backbone and training protocol: \textbf{(1) No-Adapt}: Processes historical frames as additional VLM input tokens without compression or adaptation mechanisms. \textbf{(2) Recurrent-Adapt}: Replaces our Memory Compression Module with an LSTM that sequentially processes past frame embeddings into fixed-size context representations while maintaining the identical backbone architecture as well as the training pipeline.} We leverage several representative metrics for evaluating the navigation performace: success rate (SR), success rate weighted by the ratio between the shortest path length and the predicted path length (SPL), \yhs{oracle success rate (OSR)}, trajectory length (TL), as well as navigation error (NE). Detailed information regarding simulation environment is extended in appendix.

\vspace{-1em}
\subsection{Analysis on Model Efficiencies}
\vspace{-0.5em}

\begin{figure}[t]
  \vspace{-1.5em}
  \centering
  \begin{subfigure}[b]{0.34\linewidth}
    \includegraphics[width=\linewidth]{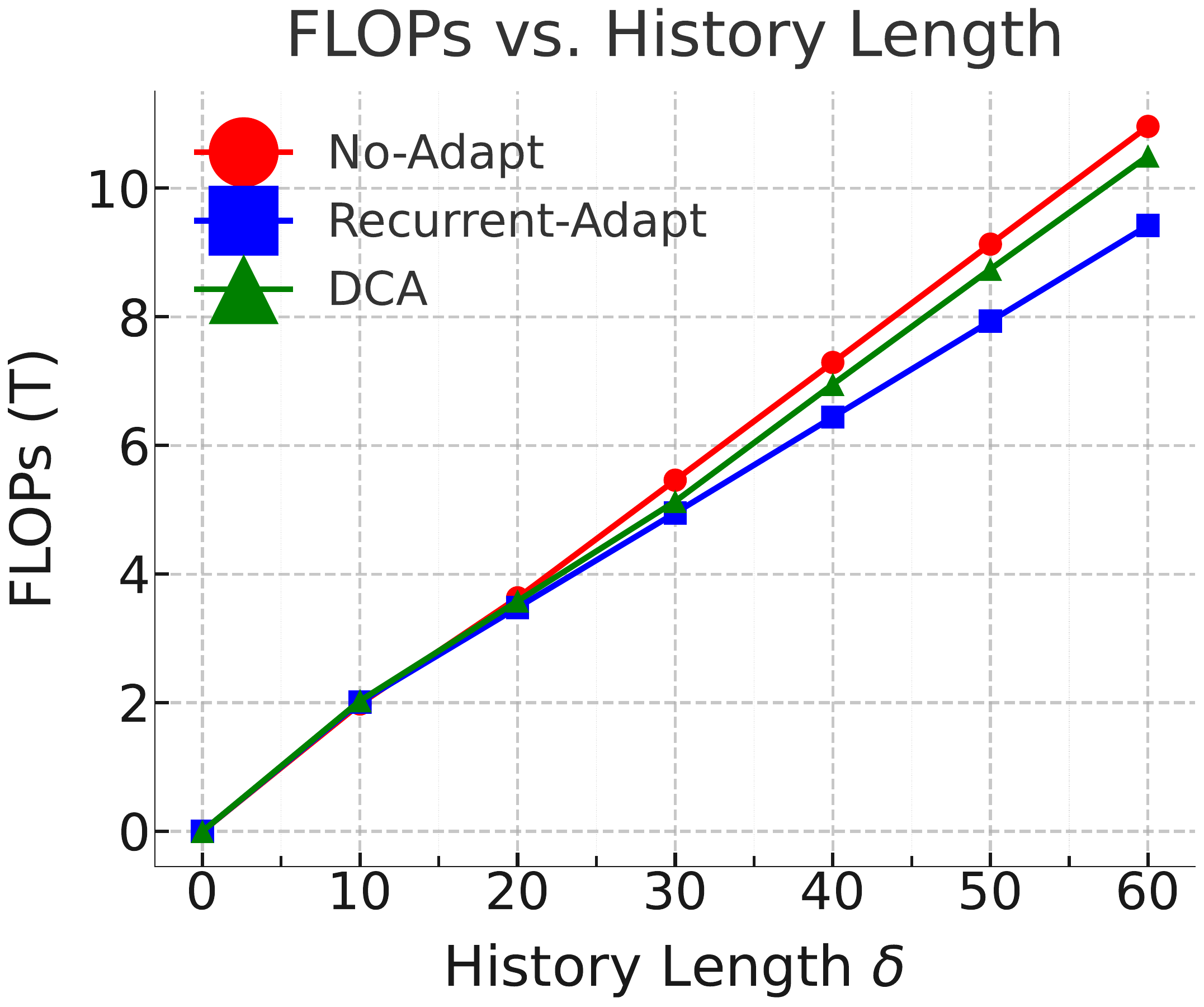}
    \caption{FLOPs vs. History Length}
    \label{fig:flops}
  \end{subfigure} \hspace{2em}
  \begin{subfigure}[b]{0.37\linewidth}
    \includegraphics[width=\linewidth]{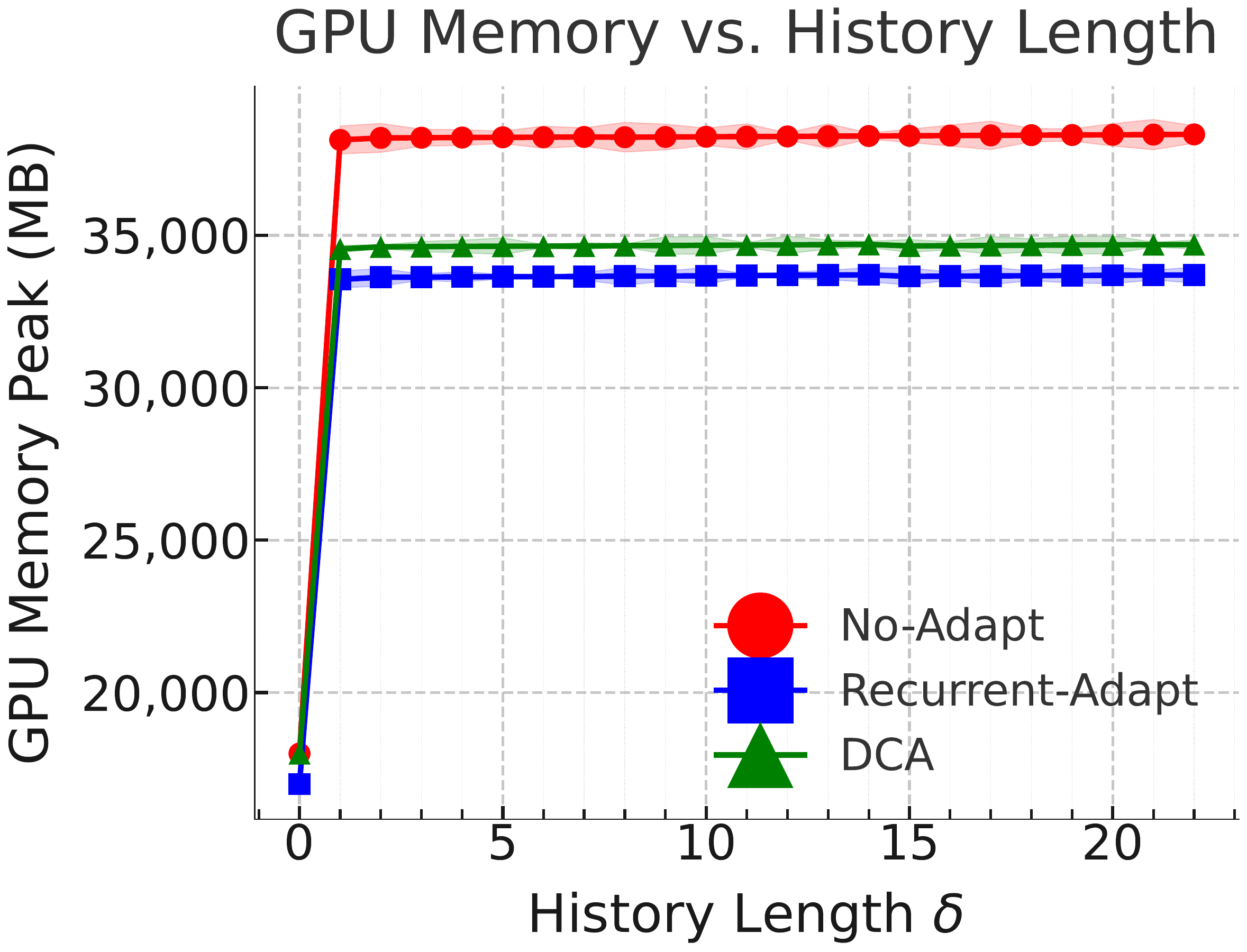}
    \caption{GPU Memory vs. History Length}
    \label{fig:memory}
  \end{subfigure}
  \caption{
  \cyl{Computational efficiency analysis of context-adaptation methods. \textbf{Left:} FLOP requirements as history length increases. \textbf{Right:} Peak GPU memory consumption across varying history lengths.}
  }
  \label{fig:flops_memory} 
  \vspace{-1.5em}
\end{figure}

\cyl{\textbf{Computational Efficiency Analysis.} Table~\ref{tab:efficiency} presents a comparison of inference throughput across methods. Our DCA approach demonstrates substantial efficiency gains compared to the No-Adapt baseline: average inference time decreases from 3.21s to 2.71s per step, FLOPs reduce from 4.77T to 4.23T, and peak GPU memory usage drops from 37.84 GB to 34.31 GB. These improvements directly result from our efficient dual-pipeline architecture that decouples historical context processing from the main VLM backbone. To analyze scalability characteristics, Fig.~\ref{fig:flops_memory} illustrates computational overhead as history length $\delta$ increases across methods. While all approaches exhibit approximately linear growth due to feed-forward network dominance independent of $\delta$, critical efficiency distinctions emerge with extended sequences. At initialization ($\delta=0$), all methods demonstrate comparable FLOP requirements. However, as history length increases, the No-Adapt baseline (red curve) exhibits the steepest computational growth. At $\delta=30$, DCA achieves over 25\% reduction in additional FLOPs relative to No-Adapt, validating our architectural efficiency claims. The Recurrent-Adapt baseline presents an interesting contrast: it demonstrates the most favorable FLOP scaling due to minimal per-timestep recurrent update costs. However, this apparent efficiency advantage comes at the expense of representational capacity, ultimately limiting navigation performance. This trade-off illustrates the fundamental challenge our method addresses: achieving both computational efficiency and representational richness for effective historical context integration. Our results demonstrate that DCA can effectively resolve this efficiency-accuracy tension through efficient architectural design rather than sacrificing either computational tractability or its temporal understanding capabilities.}

\textbf{Memory Efficiency.} Fig.~\ref{fig:flops_memory}~(b) presents GPU memory consumption patterns during training as history length $\delta$ varies. All methods exhibit modest memory growth with increasing $\delta$, as model weights dominate total memory usage. At baseline ($\delta=0$), Recurrent-Adapt shows the lowest memory requirements, while No-Adapt and DCA present nearly identical consumption. This similarity confirms that our DCA module introduces minimal architectural overhead. However, a significant efficiency gap emerges for $\delta \geq 1$: DCA consistently uses approximately 30\% less memory than No-Adapt. This reduction directly results from our compressed context representation strategy, which processes fixed-size memory vectors rather than expanding token sequences. The memory efficiency advantage becomes increasingly pronounced with longer histories, demonstrating the practical benefits of our compression-based approach for resource-constrained deployment scenarios.


\begin{wraptable}[17]{r}{0.66\textwidth} 
  \centering
  \vspace{-1.8em} 
  \caption{Evaluations on VLN-CE R2R Val-Unseen. $^*$: Methods use high-level action space. $\dag$: Methods apply the waypoint predictor proposed in~\cite{hong2022bridging}. $\ddag$: Methods use extra visual data.}
  \rowcolors{2}{cyan!15}{white} 
  \label{tab:vlnce}
  \setlength{\tabcolsep}{0.6mm}
  \resizebox{0.66\textwidth}{!}{
    \begin{tabular}{c|l|cccc|ccccc}
      \Xhline{1.2pt} 
      \multirow{2}{*}{\#} & \multirow{2}{*}{\textbf{Method}}  
        & \multicolumn{4}{c|}{\textbf{Observation}} 
        & \multicolumn{5}{c}{\textbf{VLN-CE R2R Val-Unseen}} \\ 
      \cline{3-11} 
        & 
        & Pan. & S.RGB & Depth & Odo. 
        & TL   & \textbf{NE}$\downarrow$ & \textbf{OS}$\uparrow$ & \textbf{SR}$\uparrow$ & \textbf{SPL}$\uparrow$ \\ 
      \Xhline{1.8pt}
      1  & HPN+DN$^*$~\cite{krantz2021waypoint}         
        & \checkmark &          & \checkmark & \checkmark 
        & 7.62  & 6.31  & 40.0 & 36.0 & 34.0 \\
      2  & CMA$^*$$\dag$~\cite{hong2022bridging}    
        & \checkmark &          & \checkmark & \checkmark 
        & 10.90 & 6.20  & 52.0 & 41.0 & 36.0 \\
      3  & RecurrentVLN$^*$$\dag$~\cite{hong2022bridging} 
        & \checkmark &          & \checkmark & \checkmark 
        & 12.23 & 5.74  & 53.0 & 44.0 & 39.0 \\
      4  & Sim2Sim$^*$~\cite{sim2sim}         
        & \checkmark &          & \checkmark & \checkmark 
        & 10.69 & 6.07  & 52.0 & 43.0 & 36.0 \\
      5  & HAMT$^*$$\dag$$\ddag$~\cite{hamt}
        & \checkmark &          & \checkmark & \checkmark 
        & --    & 4.80  & --   & 55.0 & 51.0 \\
      6  & LAW~\cite{law}            
        &            & \checkmark & \checkmark & \checkmark 
        & 8.89  & 6.83  & 44.0 & 35.0 & 31.0 \\
      7  & Seq2Seq~\cite{krantz2020beyond}         
        &            & \checkmark & \checkmark &             
        & 9.30  & 7.77  & 37.0 & 25.0 & 22.0 \\
      8  & CMA~\cite{krantz2020beyond}             
        &            & \checkmark & \checkmark &             
        & 8.64  & 7.37  & 40.0 & 32.0 & 30.0 \\ 
      \hline
      9  & NaVid~\cite{zhang2024navid}                           
        &            & \checkmark &           &             
        & 7.63  & 5.47  & 49.1 & 37.4 & 35.9\\
      10 & NaVid-IL~\cite{zhang2024navid}                           
        &            & \checkmark &           &             
        & --    & 7.10  & 20.6 & 14.4 & 12.4 \\
      \hline
      11 & RGB-Seq2Seq~\cite{krantz2020beyond}                             
        &            & \checkmark &           &             
        & 4.86  & 10.10 & 8.10 & 0.00 & 0.00 \\
      12 & RGB-CMA~\cite{krantz2020beyond}                               
        &            & \checkmark &           &             
        & 6.28  & 9.55  & 10.80 & 5.00 & 4.43 \\
      13 & DCA (No-Adapt)                           
        &            & \checkmark &           &             
        & 3.91  & 7.12  & 8.86 & 7.23 & 7.00 \\
      14 & DCA (Recurrent-Adapt)                           
        &            & \checkmark &           &             
        & 8.44  & 9.56  & 7.14 & 6.59 & 5.44 \\
      15 & \textbf{DCA} (Ours)                          
        &            & \checkmark &           &             
        & 6.73  & \textbf{6.77}  & \textbf{25.3} & \textbf{13.7} & \textbf{12.9} \\
      \Xhline{1.8pt}
    \end{tabular}
  }
  \vspace{-1.0\baselineskip}
\end{wraptable}

\begin{figure}[t] 
\centering

\includegraphics[width=0.9\textwidth]{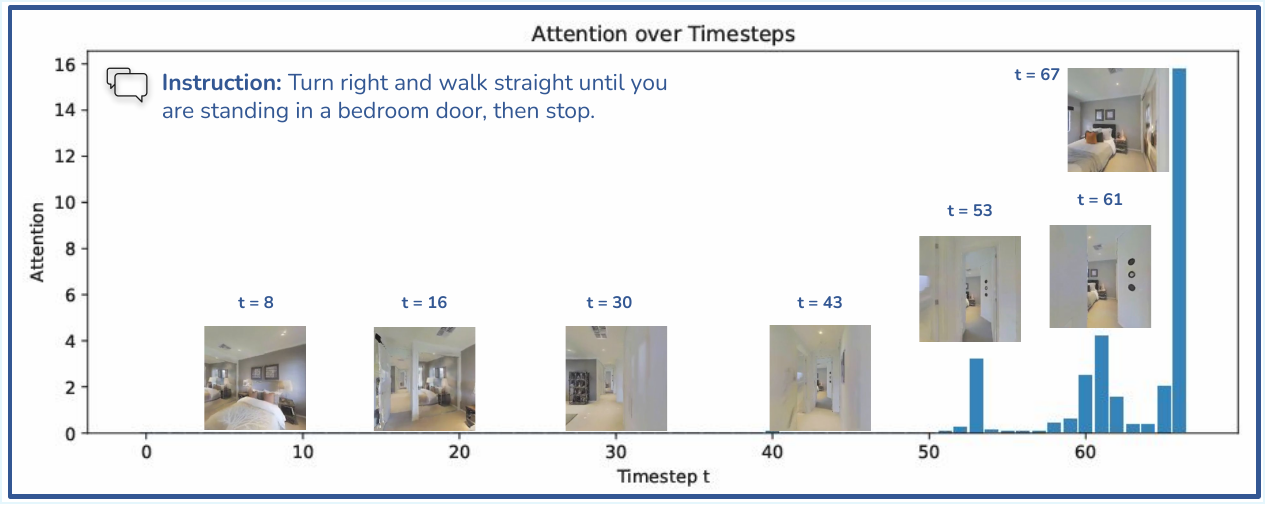}
\caption{Averaged attentions of the \emph{Memory Compression Module} across navigation timesteps for unseen evaluation episode 37 in VLN-CE, with the corresponding visual observations indicated.
}\vspace{-1.5em}
\label{atten_bar}

\end{figure}\vspace{-1.5em}

\subsection{Evaluations on \cyl{VLN} Performance}
\vspace{-0.5em}

Table~\ref{tab:vlnce} presents navigation performance on VLN-CE R2R Val-Unseen split. Methods are organized by input modality: approaches using additional sensors beyond RGB (\#1-\#8) appear above the first horizontal line, while RGB-only methods (\#9-\#15) are grouped below. NaVid variants receive separate categorization due to auxiliary co-training protocols. DCA shows substantial performance improvements under the low-level action VLN-CE framework. Compared to recurrent baselines RGB-Seq2Seq and RGB-CMA, DCA achieves relative success rate improvements of 13.7\% and 8.7\%, respectively. Against Recurrent-Adapt, which shares our backbone and adaptation framework, DCA delivers 7.11\% SR improvement, validating dynamic compression effectiveness over recurrent approaches. DCA outperforms concatenation-based approaches: it surpasses No-Adapt by 6.47\% in SR while matching NaVid-IL performance despite using a smaller backbone (3B vs. 7B parameters) and standard training rather than auxiliary co-training. The competitive Oracle Success (OS) Rate demonstrates effective instruction comprehension. These results establish DCA's superior efficiency-performance trade-offs.


\begin{figure}[t]
\vspace{-1em}
  \centering
  \begin{subfigure}[b]{0.33\textwidth}
    \includegraphics[width=\linewidth]{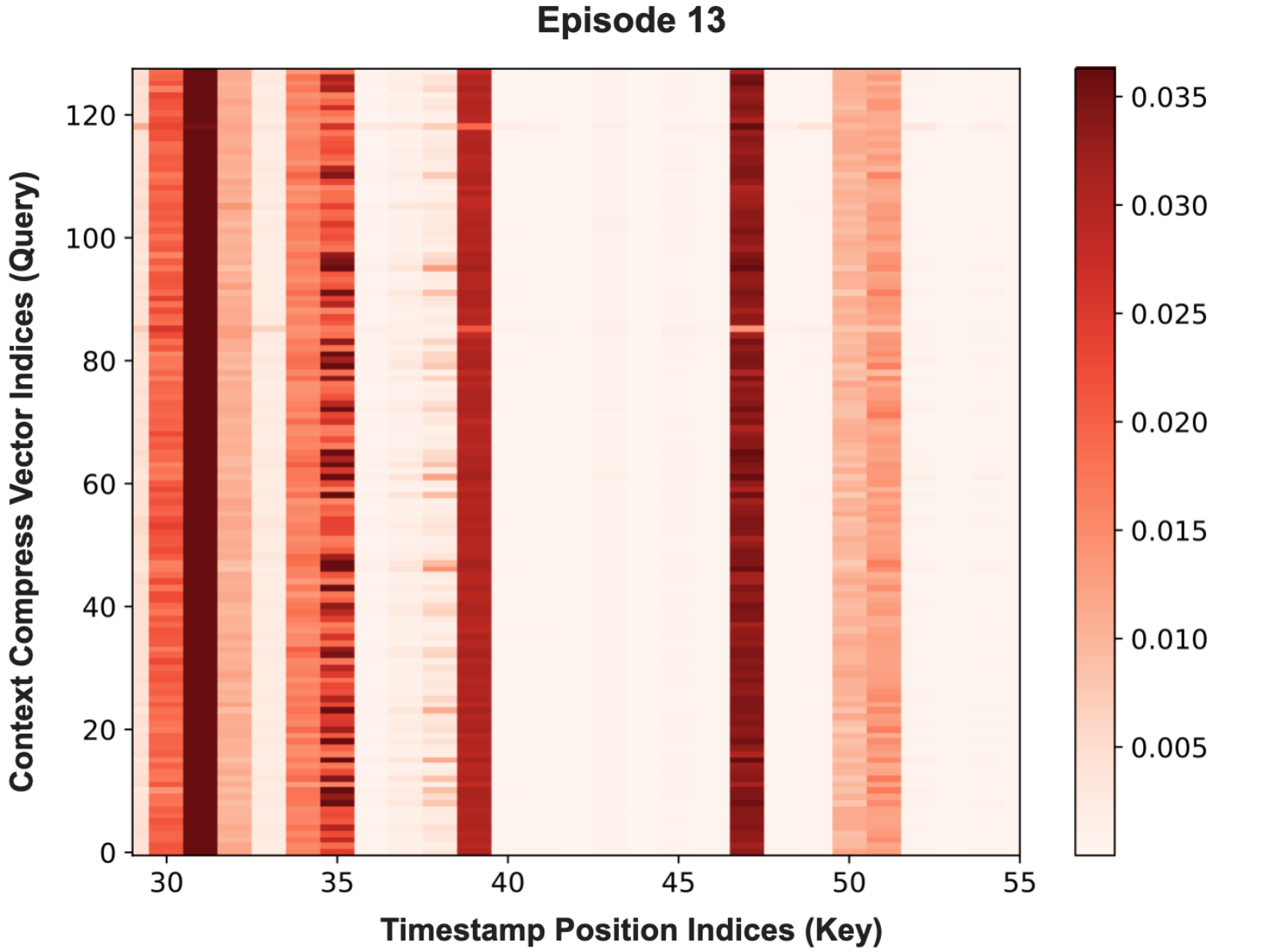}
    \label{fig:sub2}
  \end{subfigure}%
  \begin{subfigure}[b]{0.33\textwidth}
    \includegraphics[width=\linewidth]{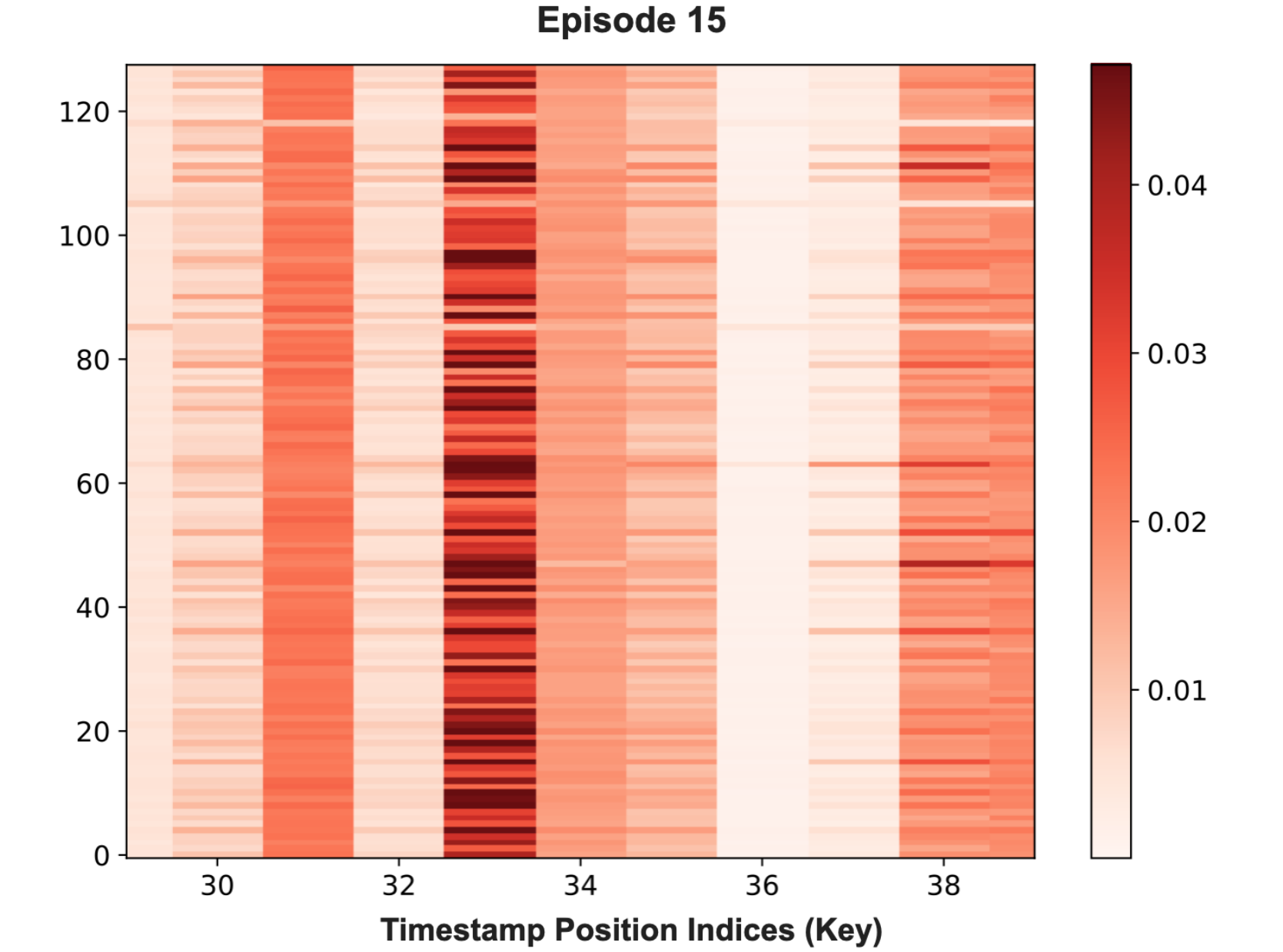}
    \label{fig:sub3}
  \end{subfigure}
  \begin{subfigure}[b]{0.33\textwidth}
    \includegraphics[width=\linewidth]{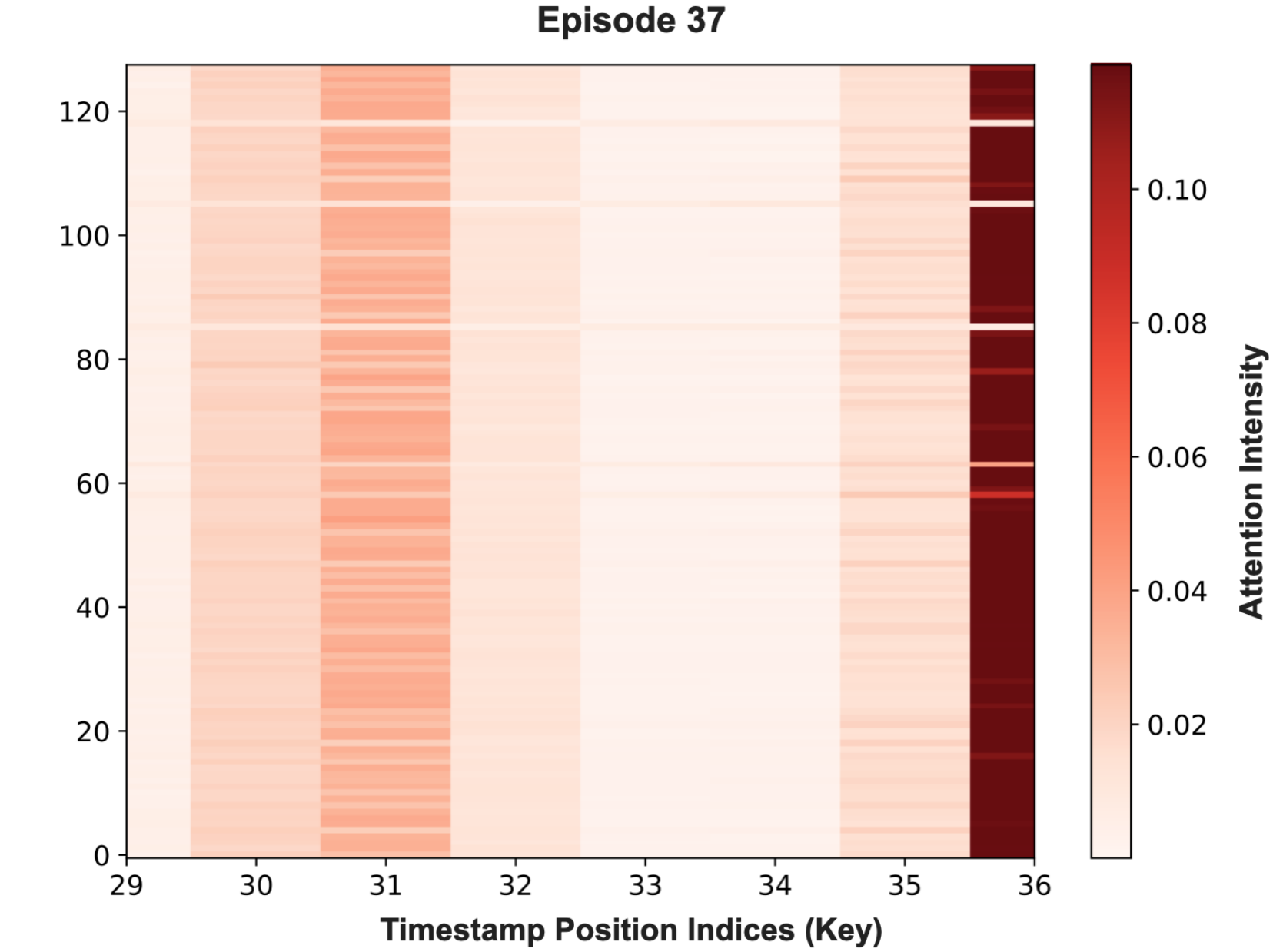}
    \label{fig:sub1}
  \end{subfigure}
  \vspace{-1.5em}
  \caption{
  \cyl{Attention heat map of the \emph{Memory Compression Module} with initial compression vectors as queries and encoded historical frames as keys, indexed by timesteps. Please note that the initial $30$ timesteps are truncated for improved visualization clarity across three representative episodes.}
  }
  \label{fig:heatmaps}
  \vspace{-1.5em}
\end{figure}

\begin{wraptable}[11]{r}{0.70\textwidth} 
\centering
\vspace{-2.4em}
\caption{The ablative investigation on feature adaptation, context compression, and compression vector length.}
\label{tab:ablation}
\vspace{-0.5em}
\scalebox{0.78}{
\setlength{\tabcolsep}{1.0mm}{
\begin{tabular}{l|lccccc}
\Xhline{1.2pt}
 & \multicolumn{6}{c}{VLN-CE R2R Val-Unseen} \\ \cline{2-7}
 & \multicolumn{1}{c|}{Type} & TL & \textbf{NE}$\downarrow$ & \textbf{OS}$\uparrow$ & \textbf{SR}$\uparrow$ & \textbf{SPL}$\uparrow$ \\ \hline

\multirow{3}{*}{\cellcolor{cyan!15}\makecell[l]{Feature\\Fusing}}
 & \multicolumn{1}{l|}{\cellcolor{cyan!15}FiLM Adapting} & \cellcolor{cyan!15}2.32 & \cellcolor{cyan!15}11.4 & \cellcolor{cyan!15}6.25 & \cellcolor{cyan!15}5.47 & \cellcolor{cyan!15}5.26 \\
 & \multicolumn{1}{l|}{$\lambda = 0.5$} & 7.34 & 7.21 & 14.6 & 10.12 & 9.66 \\
 & \multicolumn{1}{l|}{\cellcolor{cyan!15}$\lambda = 0.8$} & \cellcolor{cyan!15}6.59 & \cellcolor{cyan!15}7.01 & \cellcolor{cyan!15}17.8 & \cellcolor{cyan!15}11.4 & \cellcolor{cyan!15}9.54 \\ \hline

\multirow{4}{*}{\cellcolor{cyan!15}\makecell[l]{Context\\Compression}}
 & \multicolumn{1}{l|}{\cellcolor{cyan!15}Instruction Attention} & \cellcolor{cyan!15}7.23 & \cellcolor{cyan!15}6.90 & \cellcolor{cyan!15}8.86 & \cellcolor{cyan!15}7.23 & \cellcolor{cyan!15}6.99 \\
 & \multicolumn{1}{l|}{$C = 24$} & 12.1 & 12.3 & 8.12 & 6.94 & 5.12 \\
 & \multicolumn{1}{l|}{\cellcolor{cyan!15}$C = 48$} & \cellcolor{cyan!15}10.5 & \cellcolor{cyan!15}11.4 & \cellcolor{cyan!15}10.27 & \cellcolor{cyan!15}6.88 & \cellcolor{cyan!15}5.64 \\
 & \multicolumn{1}{l|}{$C = 64$} & 8.30 & 8.64 & 17.6 & 9.23 & 8.77 \\ \hline

\cellcolor{green!20}\textbf{Ours} 
 & \multicolumn{1}{l|}{\cellcolor{green!20}Full Setting} 
 & \cellcolor{green!20}6.73 & \cellcolor{green!20}\textbf{6.77} & \cellcolor{green!20}\textbf{25.3} & \cellcolor{green!20}\textbf{13.7} & \cellcolor{green!20}\textbf{12.9} \\ 
\Xhline{1.5pt}
\end{tabular}
}
}
\vspace{-1em}
\end{wraptable}

\vspace{-1em}
\subsection{Visualizations of Dynamic Context Compression}
\vspace{-0.5em}

\cyl{To understand our compression efficiency mechanisms, we analyze attention patterns within the \emph{Memory Compression Module} to identify which historical frames contribute most significantly to navigation decisions. This reveals how our method achieves computational efficiency through selective temporal prioritization rather than uniform compression.  Fig.~\ref{atten_bar} presents attention visualization for our compression module. Following Eq.~(\ref{eq:compressed memory vector}), we compute per-head attention scores $S_{\mathrm{cps}}$ across the trajectory at final navigation step $T=68$. These scores weight frame features from $t=0$ to $t=67$ during compression.

Aggregating scores across heads produces a per-timestep attention profile demonstrating temporal selection priorities. The analysis reveals selective focus on semantically relevant observations, validating our compression approach. Early observations without visible targets (bedroom door) receive negligible attention weights, reflecting limited utility for decision-making at $T=68$. Conversely, frames containing critical visual cues exhibit pronounced attention peaks: the target door at $t=53$ and $t=61$, and bedroom interior at $t=67$ show substantially elevated weights. Fig.~\ref{fig:heatmaps} confirms concentrated focus on later frames where goal locations become visible. These patterns confirm our method's ability to identify and prioritize critical contextual features while efficiently discarding temporally irrelevant information, achieving efficiency through intelligent temporal filtering rather than indiscriminate reduction. Additional analyses appear in Appendix.}

\vspace{-0.5em}

\subsection{Ablation Studies} 
\vspace{-0.5em}

Table~\ref{tab:ablation} analyzes design choices of our method. We compare our default context injection via Eq.~(\ref{eq:context-enhanced output}), which adds compressed context with learnable coefficient $\lambda$, against FiLM-based fusion applying $z_{k+1} \leftarrow z_{k+1} + (\alpha\,z_{k+1}^{\text{context}} + \beta)$ with zero-initialized parameters $\alpha$ and $\beta$ following~\cite{kim2025fine}. FiLM fusion shows substantial SR and SPL drops because additional scaling parameters hinder stable context integration. Varying $\lambda$ ($0.5$, $0.8$) shows larger values consistently improve success and SPL, confirming the importance of weighted historical context. For compression designs, we augment the compression module with cross-attention over instruction embeddings, hypothesizing context relevance correlates with instruction semantics. This variant underperforms direct compression due to data quality issues in R2R where instructions and trajectories are misaligned. Additionally, we examine the effect of memory capacity by varying memory tokens $C$ ($24$, $48$, $64$). Performance improvements correlate with increased $C$ values. This finding indicates that greater capacity captures richer temporal patterns, though excessive increases risk overfitting.

\vspace{-1.0em}

\section{Conclusion} \label{Sec:Con}
\vspace{-0.5em}

\cyl{In this study, we introduced DCA, a lightweight framework that efficiently integrates historical context into pretrained VLMs without inflating input token lengths. Our proposed approach employed a Memory Compression Module to distill past frame embeddings into fixed-size learnable memory vectors and a Memory Integration Module to adapt these compressed representations into each Transformer layer. This design preserved the pretrained VLM architecture while achieving linear scaling with extended context lengths. Our extensive evaluations on downstream VLN tasks demonstrated that DCA can achieve superior efficiency-performance trade-offs compared to existing approaches.}




\vspace{-1.0em}
\section{Acknowledgement}
\vspace{-0.5em}

The authors gratefully acknowledge the support from the National Science and Technology Council (NSTC) in Taiwan under grant numbers NSTC 114-2221-E-002-069-MY3, NSTC 113-2221-E-002-212-MY3, and NSTC 114-2218-E-A49-026. This research was also supported by the NVIDIA Academic Grant Program. The authors would also like to express their appreciation for the donation of the GPUs from NVIDIA Corporation and NVIDIA AI Technology Center (NVAITC) used in this work. Furthermore, the authors extend their gratitude to the National Center for High-Performance Computing (NCHC) for providing computational and storage resources. The authors also thank the NVIDIA Taipei-1 supercomputer for providing essential computing resources.

%
%
%
\bibliographystyle{splncs04}
\bibliography{bibs}
%




\end{document}